# How critically can an AI think? A framework for evaluating the quality of thinking of generative artificial intelligence.

*Luke Zaphir, Jason M. Lodge, Jacinta Lisec, Dom McGrath, Hassan Khosravi*

*The University of Queensland, Australia*

## Abstract

Generative AI such as those with large language models (LLMs) have created opportunities for innovative assessment design practices. Due to recent technological developments, there is a need to know the limits and capabilities of generative AI in terms of simulating cognitive skills. Assessing students' critical thinking skills has been a feature of assessment for time immemorial, but the demands of digital assessment create unique challenges for equity, academic integrity and assessment authorship. Educators need a framework for determining their assessments' vulnerability to generative AI to inform assessment design practices. This paper presents a framework that explores the capabilities of the LLM ChatGPT4 application, which is the current education industry benchmark. This paper presents the Mapping of questions, AI vulnerability testing, Grading, Evaluation (MAGE) framework to methodically critique their assessments within their own disciplinary contexts. This critique will provide specific and targeted indications of their questions vulnerabilities in terms of the critical thinking skills. This can go on to form the basis of assessment design for their tasks.

*Practitioner notes:*

*What is already know about this topic:*

- Digital technologies have supported students completing assessments for decades
- Generative AI can pose concerns for determining authorship and integrity of assessment
- Several approaches to addressing Generative AI have been offered, including invigilation, designing around the technology, and rethinking the task itself
- There are methods for measuring the quality of thinking but many are focused on specific cognitive skills such as argumentation
- The skill of the Generative AI user can produce significant differences in the quality of the output

*What this paper adds*

- A viable framework (MAGE) for precisely evaluating the vulnerabilities of assessment tasks through a critical thinking perspective.
- Specific language that can be used to describe the quality of thinking from contemporary critical thinking theory and practice
- Insight into the contemporary abilities of ChatGPT4's ability to emulate critical thinking skills and values

*Implications for practice and/or policy*

- **Enhanced Assessment Design**: Recognizing the capabilities of generative AI will be pivotal for educators. It will direct them towards crafting assessments that deeply probe students' comprehension and analytical skills, rather than mere information recall which can be easily manipulated with AI assistance.
- **Emphasis on Authentic Learning**: In the age of readily available AI tools, there's an increased urgency for educators to stress on experiential and project-based learning. Such methods not only deter AI-driven shortcuts but also nurture genuine understanding and application of knowledge in real-world scenarios.

# Introduction

Since late 2022, generative artificial intelligence applications using large language models (LLM) have become internationally and ubiquitously available. With generative AI becoming ubiquitous, contemporary approaches to teaching, learning and assessment have been disrupted presenting both major opportunities for novel learning and threats to academic integrity. With the exception of invigilated tasks, the authorship of many assessment tasks is potentially unknowable, raising questions about whether unsupervised assessment can fulfil educational purposes. Authorship is potentially now in question when a non-invigilated assessment can be completed. Increased precision is needed for assessment design to ensure that assessment artefacts created by students demonstrate the skills and competencies they are required to.

Using generative AI to produce assessment artefacts provides students with the potential to fulfill assessment task requirements without the requisite cognitive skills. This differs from other types of potential academic misconduct such as contract cheating, where a student may pay to outsource their assessment to be completed by another person or agency (Clarke & Lancaster, 2006). When students use generative AI to produce assessment artefacts, the line between student and machine is far murkier. A student may engage in a dialogue with ChatGPT, refining their text through its digital tutelage. Alternately, they may create highly sophisticated instructions for the generative AI (prompt engineering), which can produce responses that are able to complete or partially complete an assessment task.

There are a number of assessment design approaches to address the issue of students using Generative AI to produce assessment artefacts, such as invigilation, design around (creating a task that is mindful of AI's limitations) and rethink (changing the task to more precisely match the learning needs of the student) (Lodge, 2023). Rethinking each assessment task is one of the few long-term solutions, though achieving this is challenging. ChatGPT4 can emulate critical thinking in numerous ways, thus requiring educators to use a method to determine how vulnerable their assessment task is to being produced by ChatGPT4 with academic success, to achieve a pass or higher. Knowing which types of critical thinking can be performed by ChatGPT4 and to what level of quality is essential to inform assessment design.

Critical thinking involves diverse skills and dispositions with methodical rigour that can only be mimicked by generative AI rather than performed by it. Beyond merely using cognitive skills, critical thinking also involves values of inquiry which can be used to judge the quality of thinking. This article describes a framework (MAGE – mapping, AI vulnerability testing, grading, evaluation) that can determine the vulnerability of an assessment task to being successfully completed by generative AI in terms of critical thinking. This framework is based on this study's two research questions, to explore potential opportunities and risks of using a critical thinking approach to design assessment.

- RQ1: What are the current capabilities of ChatGPT4 to perform cognitive skills across critical thinking domains? This question aims to understand how effectively ChatGPT can interpret and produce the requirements of a written assessment task prompt, involving one or more cognitive verbs. ChatGPT may be able to analyse a text but how effectively can it analyse? This question aims to evaluate these aspects of critical thinking.
- RQ2: What criteria from the values of inquiry can be evidenced as most challenging to assess against criteria descriptors for a pass grade or higher? This question provides a mechanism for measuring the quality of critical thinking skills required in a written assessment task. This question aims to at assist educators to design questions for written assessment that are resilient to Generative AI producing an assessment artefact with academic success of a pass or higher.

The rest of this paper is organised as follows. First, an explanation of how critical thinking can be measured. Next an outline of the framework functions, including mapping the task to critical thinking skills, testing for vulnerabilities, grading, and evaluating to indicate specific vulnerabilities. This is followed by a case study demonstrating the framework's use when applied to a higher education assessment task. A brief overview of the current state of critical thinking vulnerabilities follows. The last section includes the limitations of the approach, along with recommendations and implications for assessment design.

## *Measuring the quality of critical thinking*

Generative AI represents an uncertain and evolving education technology due to its continuously updated knowledge, data and capabilities (Giannini, 2023), and how people are responding to these updates with a myriad of new uses (E. R. Mollick & Mollick, 2022). What do we mean by capabilities though? Much of what we do as educators requires the development of cognitive skills, but this does not quite capture the breadth of learning experiences. What is needed is a method of critiquing/evaluating which components of thinking that LLM Generative AI can and cannot emulate effectively, and what level of threat this presents to the learning objectives of an assessment task.

A framework for critiquing an assessment task's vulnerability to generative AI would need to accomplish several goals:

1. Clarify the critical thinking task to be evaluated
2. Describe a method that evaluates the quality of critical thinking against grade descriptors
3. Indicate the strengths and weaknesses of the assessment question

For the first two tasks, a definition of critical thinking is needed. There has been an abundance of taxonomies for critical thinking that have been used for assessment, and there is no consistent definition of skills or competencies when it comes to thinking. Pelligrino & Hilton (2012) argue that there is considerable overlap between cognitive skills, competencies and personality traits; taxonomies designed to quantify these qualities often need to be adjusted or reconceptualised to suit the context in order to be a transferrable skill.

There has been ongoing debate around the methods for measuring these skills and competencies are not agreed on by experts. The 1990 Delphi report agreed that a number of skills were fundamental to critical thinking (i.e. interpretation, analysis, evaluation, inference, explanation, self-regulation) but did not have a method of measuring these (Facion, 1990). Several validated mechanisms have been created, such as the California Critical Thinking Skills test (CCTS) and the Watson-Glaser Critical Thinking test (El Hassan & Madhum, 2007), but these explore de-contextualised skills. Specific critical thinking measurement methods have been created but these are usually contextualised within a discipline (Lawson, Jordan-Fleming, & Bodle, 2015), or as argumentation skills (Wulandari & Hindrayani, 2021). Many of these tests cannot be applied directly to LLM Generative AI in a way that is useful to determining a task's vulnerability to being generated and completed by AI.

A possible method of measuring critical thinking is to examine the cognitive skills and qualities of those thoughts. Ellerton (2022) argued that critical thinking can be categorised into several domains. The first, skills of inquiry, are the cognitive skills that are being performed. Clarity is needed on what cognitive skill the AI is being asked to perform and how it is being asked specifically. Second is values of inquiry; language that can be used to evaluate the quality of the cognitive skill. Ellerton & Kelly argued (2022, p. 17) argued that there are many epistemic values that are important to skilled reasoning such as accuracy, precision, coherence, relevance[1]. This framework uses these types of values as discrete criteria for measuring the quality of thinking, but does not impose restrictions on the number of types of values that must be included for each test. In this way, the framework can provide a consistent method for educators to evaluate the quality of students' critical thinking in written tasks, in ways that are not possible or easy to be emulated by an LLM generative AI.

This standard is useful as provides language for both what is being evaluated (e.g. a cognitive skill such as analysis) and language for evaluating (e.g. accuracy, depth, relevance). These values of inquiry were originally based on the Intellectual Standards created by Paul & Elder (2013), who argued that these were conceptualisations of the possible weaknesses and strengths of thinking. Using this, it becomes feasible to create language to evaluate the relative qualities of thinking in a LLM like ChatGPT. This framework may then be used as an indicator of whether an assessment task needs to change and how.

## The MAGE Framework

### Mapping, AI vulnerability testing, Grading, Evaluation

The MAGE framework is designed for educators to follow four-steps to methodically evaluate an assessment question in terms of its vulnerability and strengths to being completed by Generative AI with academic success to achieve a pass grade or higher. The user must determine the version of this question they are going to use as a prompt to evaluate, as this can take the form of the original question, a version which has been refined to describe the cognitive skill involved, or a prompt engineered version which greatly elaborates on the instructions for the AI. This question would then be inputted into the Generative AI and the response graded against criteria (based on eight values of inquiry outlined by Ellerton previously. This grading would then be evaluated the specific vulnerabilities and strengths of the question.

---

[1] The values selected for the case studies in this paper are clarity, accuracy, precision, depth, breadth, coherence, relevance and significance. There are other values but these eight are common to many disciplines. Other values such as cogency (persuasiveness), simplicity, reproducibility etc are less ubiquitous.

*Step 1: Mapping*

Prior to testing, the question is mapped to its specific cognitive skill. Appropriate values of inquiry are then selected as the criteria for grading this skill. Only one cognitive skill will be tested at a time, though multiple values can be used as criteria.

Example 1: Question for an essay assessment task: *What caused the sinking of the Titanic?*

Table 1: Example question mapping

| Cognitive skill | Possible cognitive verbs | Value(s) of inquiry |
| --- | --- | --- |
| Explanation | Explain, describe, elaborate, clarify | Accuracy, Precision |

The value(s) of inquiry selected must also be contextualised; critical thinking is always applied within a specific knowledge and skill area. A value of inquiry must reflect the needs of that context.

Table 2: Example grade descriptors

| Criteria | High achievement | Pass | Fail |
| --- | --- | --- | --- |
| Accuracy | Describes information and arguments that are factually correct related to the Titanic's sinking. | Describes mostly accurate information with minor errors, generally related to the Titanic's sinking. | Describes with multiple inaccuracies related to the Titanic's sinking |
| Precision | Describes with exceptional quality of detail and specificity directly related to the Titanic's sinking. | Describes some quality detail with minor vagueness, mostly specific to the Titanic's sinking. | Describes without adequate detail and specificity; the content is vague, ambiguous, or not directly related to the Titanic's sinking. |

*Step 2: AI Vulnerability testing*

There can be three possible versions of the question to be tested. First, the original unchanged version. Second, a minimally adapted question which has been cognitively audited by specifying the type of cognitive skill that the question is asking the student to perform, and then adding a specific cognitive verb for this skill). Third, a prompt engineered question, meaning that the input into the Generative AI has been designed to both follow contemporary guidelines[2] (Microsoft, 2023; E. Mollick & Mollick, 2023) for prompt enhancement, as well as include specific aspects of the criteria. This is to ensure that meaning is communicated fully and the generative AI is focused on the desired value of inquiry. When inputting each question into the Generative AI, the response should be regenerated several times to ensure consistency.

Table 3: Example questions with alternate versions

| Original question | Minimally adapted | Prompt engineered |
| --- | --- | --- |
| *What caused the sinking of the Titanic?* | *Describe the causes of the sinking of the Titanic.* | *You are a tutor who helps students understand concepts by explaining in highly accurate and precise, specific detail. Your answers should be full paragraphs using language appropriate to an adult reading level. Each different cause should be in its own paragraph, with each paragraph structured in the TEEL style without signposting. Your response should be in essay style* |

---

[2] Prompt engineering may become less of a factor over time. Generative AI such as Claude (Anthropic, 2023) are more intuitive to the user and are more predictive. Conversely, deep generative models have the ability to have specific objectives and focus on particular skills and tasks (Chu, 2023), also diminishing the need for prompt engineering. It is unclear what the long-term ramifications of these developments will be.

*Step 3: Grading*

Each of the test responses must be graded against the grade descriptors. Grading should be conducted by markers with appropriate levels of subject matter expertise. Each grade should have a clear explanation for receiving that level of achievement.

Table 4: Example grading[3]

| Question variant | Level of achievement (Accuracy) | Level of achievement (Precision) |
|---|---|---|
| Original question | High | Fail |
| Minimally adapted | High | Fail |
| Prompt engineered | High | Pass |

*Example grading analysis:*

*All responses demonstrated a high level of accuracy about the causes of the sinking of the Titanic. However, Both the original question and the minimally adapted question provided minimal levels of detail about sinking of the Titanic. The prompt engineering was less vague and went into passable detail about the sinking, elaborating on each of these causes.*

*Step 4: Evaluation*

Determining the vulnerability of the question requires analysing several factors, including how easily the Generative AI fulfilled the task with each variant when measured against the value of inquiry. This vulnerability can said to match the following levels:

Table 5: Example vulnerability values

| Level of vulnerability by value[4] | Description |
|---|---|
| Minor | The task cannot be easily completed even with extensive prompt engineering |
| Low | The task fails can only be pa1ssed with extensive prompt engineering |
| Moderate | The task can be passed with minimal adaptation or receives high grades with prompt engineering |
| Major | The task can receive a high achievement in its original form or with minimal adaptation |

Table 6: Example vulnerability outcomes

| Value of inquiry | Accuracy | Precision |
|---|---|---|
| Vulnerability level | Major | Low |

*The cognitive task has a major level of vulnerability in terms of the value of accuracy, producing responses that explore all of the factors that caused the sinking of the Titanic. The task has low vulnerability in terms of Generative AI using the value of precision however. There were some details included in the original and minimally adapted responses, but these were somewhat vague. Prompt engineering improves the quality of the response to a passable level.*

---

[3] Responses from ChatGPT4 can be read in full in Appendix 1
[4] "No vulnerability" has been excluded from this list as this would indicate the question cannot be fulfilled by an LLM. This would constitute 'design around' rather than 'rethink' as an approach to Generative AI.

## *Case study*

*Input*
Assessment task to be evaluated: Undergraduate Biology Reflective Writing task

*"The bush holds different cultural significance for different people. Choose one piece from the exhibition by an artist living in Australia. What relationship to the desert does the artwork portray? How does this vary from your own cultural attachment to the bush? (500 words or less)"*

Table 7: Case study mapping

| *Step 1 Question mapping* | |
|---|---|
| **Cognitive audit** (map question to cognitive skill) | **Values audit** (Select appropriate values of inquiry for testing) |
| Reflection – cognitive verb 'reflect' | Relevance  Significance  Depth  Coherence |

| *Grade descriptors* | | | |
|---|---|---|---|
| Criteria | High | Pass | Fail |
| Relevance | Relates experiences to broader contexts effectively. | Partially relates experiences to broader contexts. | Doesn't relate experiences to contexts. |
| Significance | Profound understanding across all contexts. | Basic understanding across contexts. | Lacks contextual significance understanding. |
| Depth | Highly insightful with evidence. | Some insights or uses evidence. | Shallow insights and lacks evidence |
| Coherence | Highly ordered, logical thought structure. | Mostly structured and cohesive. | Incoherent, inconsistent ideas. |

Table 8: Case study AI vulnerability testing

| *Step 2: AI vulnerability testing* (Test three versions of the questions: the original, minimally adapted and prompt engineered against Generative AI | | |
|---|---|---|
| Original question | Minimally adapted | Prompt engineered |
| *The bush holds different cultural significance for different people. Choose one piece from the exhibition by an artist living in Australia. What relationship to the desert does the artwork portray? How does this vary from your own cultural attachment to the bush? (500 words or less)* | *Write a 500 word reflective writing piece on the following topic: "The bush holds different cultural significance for different people. Choose one piece from the exhibition by an artist living in Australia. What relationship to the desert does the artwork portray? How does this vary from your own cultural attachment to the bush?"* | *You are an art student who has lived their whole life in Alice Springs, Australia. You feel a deep connection to the land as a Mparntwe area. You speak Arrente as a language and are a passionate about maintaining the cultural heritage of your people.  Write a 500 word reflection on the cultural significance of the artist Josie Petrick Kemarre's work "Bush Plum Dreaming" dot-work painting, focusing on how the work depicts an edible berry bush - one of the few food sources in a vast, Australian desert. Include a reflection on the kinship and pride you have to the author, and gratitude towards indigenous art that is meaningful to your identity.* |

Table 9: Case study grading

| | Step 3: Grading[5] | | | |
| :---: | :---: | :---: | :---: | :---: |
| | (Grade the response against the rubric) | | | |
| | Level of achievement | | | |
| Question variant | Significance | Relevance | Depth | Coherence |
| Original question | Fail | Fail | Pass | Pass |
| Minimally adapted | Fail | Fail | Pass | Pass |
| Prompt engineered | High | High | High | High |

Table 10: Case study evaluation

| Step 4: Evaluation | |
| :---: | :--- |
| **Vulnerability Rubric** | |
| (Develop a vulnerability rubric) | |
| Level | Description |
| Minor | The task cannot be easily completed even with extensive prompt engineering |
| Low | The task inconsistently passes with minimal adaptation or receives a pass or higher with prompt engineering |
| Moderate | The task passes with minimal adaptation and receives high achievement with prompt engineering |
| Major | The task can receive a high achievement in its original form |

| **Vulnerability outcome** | | | | |
| :---: | :---: | :---: | :---: | :---: |
| (Determine the level of vulnerability for each selected value) | | | | |
| Value of inquiry | Significance | Relevance | Depth | Coherence |
| Vulnerability level | Low | Low | Moderate | Moderate |

## *Case study discussion:*

The original question and minimally adapted questions produced reflective writing pieces of a similar quality. Both were somewhat logical in their structures with sufficient depth to demonstrate reflective thought. However, in both cases there was no discussion of an existing artist living in Australia – both produced hallucinations of both author and artwork. As a result, they both failed to relate experiences to context as well as lacking contextual significance. Prompt engineering created a vastly superior achievement. This required the addition of a real artist and artwork, as well as the cultural context of the author and a deliberate expression of their thoughts and feelings. ChatGPT produced an output that demonstrated high levels of depth, relevance, significance and coherence as a result of this extensive prompting.

**Significance** – in this particular case, significance required reflecting on the importance of various aspects of the artwork when compared to one's own experience ("across contexts"). In the original question, the response notably admitted to being an AI and spoke to the potential experiences others had, failing this criterion. The minimally adapted response was better, though not enough to pass for a different reason – hallucinations. A response that would reach 'pass' must speak to real experiences and this was obviously not doing so. Prompt engineering produced a high quality response due to two reasons: the use of real artwork and a real artist, and the inclusion of which emotional and intellectual aspects were important as part of the prompt instruction.

**Relevance** – this criterion requires that the reflection connects the artwork to the experiences of the student. Without experiences to connect to, the AI cannot meaningfully complete this task. Prompt engineering is required for the response to demonstrate relevance. Once these experiences are made explicit in the prompt, the Generative AI can produce high quality responses. This task receives a low vulnerability for relevance the same reason as significance – with prompt engineering it has highly relevant responses, but fails without them.

**Depth** – The original question and minimally adapted versions both demonstrated a passable level of depth in the reflection, providing some detail into the fictitious artworks and experiences but with some superficial components too.

---

[5] Responses from ChatGPT4 can be read in full in Appendix 2

Prompt engineering boosted these into high levels of achievement, thus receiving a moderate vulnerability rating for depth.

**Coherence** – the reflection is set out in a structurally consistent manner, but without a high level of cohesion. There is no overall logical purpose or through-line of reflection in the original or minimally adapted versions. With prompt engineering the response becomes highly coherent, as the prompt creates the logical structure for the AI to use. This indicates a moderate vulnerability rating for coherence.

*Vulnerability conclusion:*
High levels of detail and prompting are needed for this task to be completed by a generative AI to a passable standard. The requirements of this task to connect a real-life artwork to the students' own experiences as well as reflect on the importance of those experiences, makes this difficult for an AI to perform. Additionally, a student may meaningfully fulfil this reflective writing task through effective prompt engineering; they would need to share their ideas, beliefs, cultural experiences and other relevant thoughts to create the prompt itself. This task overall[6] has a low vulnerability rating. No changes are needed to the task.

## The State of the Art

The quality and capabilities of Generative AI are certain to improve over time. However, this approach was applied to the six mentioned cognitive skills[7] (reflection, interpretation, justification, evaluation, analysis, explanation) and while the results of these vulnerability tests are transient, they can be considered broadly indicative of ChatGPT4's capabilities as of October 2023. The following table represents the application of the MAGE framework across multiple disciplines.

|  | Reflection | Interpretation | Justification | Evaluation | Analysis | Explanation |
|---|---|---|---|---|---|---|
| Clarity | Mod. | Major | Low | Mod. | Low | Major |
| Accuracy | Mod. | Major | Low | Major | Mod. | Major |
| Precision | Mod. | Low | Low | Major | Major | Mod. |
| Breadth | Low | Low | Major | Major | Major | Mod. |
| Depth | Mod. | Low | Low | Low | Low | Major |
| Relevance | Low | Low | Major | Major | Major | Low |
| Significance | Low | Low | Low | Low | Low | Low |

While broad conclusions cannot be drawn from the results of a small sample size, it can be indicative of some vulnerabilities that already exist. The following values can be said to be vulnerable by task:

| Cognitive Skill | Major vulnerabilities |
|---|---|
| Interpretation | Clarity, Accuracy |
| Justification | Breadth, Relevance, Coherence |
| Evaluation | Accuracy, Precision, Breadth, Relevance |
| Analysis | Precision, Breadth, Relevance |
| Explanation | Clarity, Accuracy, Depth |

This has potential implications for assessment design; question authors may want to re-target the value being assessed on these skills (e.g. to shift from 'interpreting with accuracy' to 'interpreting with precision'), or de-emphasise their weighting when apportioning marks. If these values are crucial for the assessment task, other approaches such as invigilation or designing around AI limitations may be more appropriate. At present, significance is the least vulnerable value of inquiry with all of its outcomes being low. In almost all cases, the AI did not know which information was important or why it was important. Once a prompt was engineered with this information, it could perform well on the 'significance' criterion.

---

[6] Creating overall ratings is not necessarily advisable as the nuances of quality of thinking can provide educators with specific guidance on the strengths and vulnerabilities of their assessment tasks.
[7] Each cognitive skill was tested using the approach twice, using four appropriate values of inquiry. These tests were conducted across a variety of disciplines.

## Limitations of the framework

### Prompt Engineering

The quality of the MAGE framework is dependent on the quality of the prompt. Prompt engineering – the process of specifying any number of conditions of the response (role, perspective, structure etc) – has enormous effects on the quality of the output. There is near limitless potential for variability in both the way prompts are written as well as the way the user is interacting with ChatGPT. In some instances, a topic or issue is known well enough by the Generative AI to produce an adequate or even excellent result in terms of the cognitive skill and value of inquiry (such as with clarity of explanation).

Alternatively a user may use ChatGPT more dialogically than this. Rather than engineering a highly sophisticated initial prompt, they may use the question as the starting point for an ongoing process of review and refinement. They may achieve a desired outcome through dozens of small prompts in response to outputs, rather than one lengthy prompt and output.

Even the question design itself can significantly alter the outputs from ChatGPT; a question deliberately or explicitly designed to evoke critical thinking in a specific manner may produce this result from generative AI. A question may imply that it is asking more due to the context and this may create worse outputs. Any person using this framework will have to review and revise their prompts to ensure that consistency is achieved to the level that they need.

### Grading scale

These assessments responses were marked against grade descriptors for high, pass and fail based on selected values of inquiry to judge the quality of critical thinking as a concession to the practicality of grading hundreds of responses. For best use of the framework, grade descriptors would need to be specific to the assessment task and discipline with precision for each level of achievement. This would avoid issues like the one above where a response may have been passable but only marginally better with prompt engineering. The grading of responses would also ideally need to be moderated between several markers to ensure consistency.

### Missing elements of critical thinking

Not all aspects of critical thinking have been included in this framework, notably metacognition and virtues of inquiry. These can influence the testing of Generative AI but cannot be easily measured. Metacognition, or thinking about thinking, requires that the person explore the source and quality of their own thoughts. Generative AI does not think and so does not have metacognition, but the way it has been programmed and its inherent biases are quasi-metacognitive factors. Virtues of inquiry are the attitudes and mental habits (and appreciation thereof) that the person has while exploring the content and applying cognitive skills; ideally students should be open-minded, honest and rigorous (amongst many other intellectual qualities). These kinds of virtues can be built into prompt engineering and have the potential to alter the quality of responses.

## Considerations for assessment design

Each step of the framework provides different opportunities for reconsideration and redesign of assessment tasks. Question mapping can be used to confirm or alter the focus of the question itself. This could involve any number of changes: range from changing the cognitive skill to re-weighting the grade descriptors to de-emphasise the more vulnerable values.

Students may have access to a wide variety of assessment resources including rubrics. It would be a relatively trivial matter for a student to look to this rubric for specific cognitive skills (e.g. explain) and the value by which it is measured (e.g. clarity) and then to create a prompt that includes "explain with a high degree of clarity…".

This framework allows for educators to determine how vulnerable their assessments are to this kind of gaming. This would not solve the immediate problem of the AI-vulnerable assessment task but would provide them with insights about where the vulnerabilities lie, and perhaps indicate directions for either altering assessment (e.g. to include activities that can't be easily generated by LLMs or to change the value of inquiry evaluated etc). Alternatively, learning activities could be developed to lean into these skills and values so that using ChatGPT would become a cost-ineffective method of cheating.

## *Recommendations*

These results present opportunities for targeting assessment practices to be resilient to generative AI completion, as well as vulnerabilities to avoid if at all possible. For example, tasks that require accurate interpretation would be easily completed by AI, whereas an assessment that required a precise reflection would be challenging at present. Educators are encouraged to replicate our process with their assessment and current models. Educators may also want to consider what is an acceptable or unacceptable level of vulnerability for their tasks. Receiving a high-level achievement with minimal prompting would be the worst-case scenario for educators and is worth noting. A task that performs well in one value but poorly could have its criteria re-weighted to de-emphasise the vulnerable value. If these vulnerabilities are known prior to deploying the task, it could be re-designed so that the question focuses on different critical thinking skills.

## Conclusion

ChatGPT and other LLMs are continuing to evolve and this constitutes an ongoing challenge to assessment design. Regardless of how intuitive these generative AI become, there will always be a need for students to develop and demonstrate critical thinking skills. The MAGE framework aims to provide educators the evidence needed to rethink their assessment. There is no one approach that will suit all; it is the versatility of the framework that will allow the method to be useful in this rapidly changing education technology environment.

## Ethics statement

The research above is exempt from ethics approval due to its theoretical, analytical, and computational nature, with no empirical data involving humans or animals collected or utilized.

## *References*

Anthropic. (2023). Introducing Claude. Retrieved from https://www.anthropic.com/index/introducing-claude

Chu, J. (2023). To excel at engineering design, generative AI must learn to innovate, study finds. Retrieved from https://news.mit.edu/2023/generative-ai-must-innovate-engineering-design-1019

Clarke, R., & Lancaster, T. (2006). *Eliminating the successor to plagiarism? Identifying the usage of contract cheating sites.* Paper presented at the Proceedings of 2nd international plagiarism conference.

El Hassan, K., & Madhum, G. (2007). Validating the Watson Glaser critical thinking appraisal. *Higher Education, 54*(3), 361-383.

Ellerton, P. (2022). The Values of Inquiry: explanations and supporting questions. Retrieved from https://critical-thinking.project.uq.edu.au/files/6580/VOI%20poster%20B2.pdf

Ellerton, P., & Kelly, R. (2022). Creativity and critical thinking. In *Education in the 21st Century: STEM, Creativity and Critical Thinking* (pp. 9-27): Springer.

Facion, P. A. (1990). Critical Thinking: A Statement of Expert Consensus For Purpose Of Education Asessment And Instruction. *California Stat University*.

Giannini, S. (2023). Reflections on generative AI and the future of education. *UNESDOC digital library*.

Lawson, T. J., Jordan-Fleming, M. K., & Bodle, J. H. (2015). Measuring psychological critical thinking: An update. *Teaching of Psychology, 42*(3), 248-253.

Lodge, J. M., Howard, S., & Broadbent, J. (2023). Assessment redesign for generative AI: A taxonomy of options and their viability. *LinkedIn Pulse*. Retrieved from https://www.linkedin.com/pulse/assessment-redesign-generative-ai-taxonomy-options-viability-lodge/

Microsoft. (2023). Prompts-for-edu. Retrieved from https://github.com/microsoft/prompts-for-edu/

Mollick, E., & Mollick, L. (2023). Assigning AI: Seven Approaches for Students, with Prompts. *arXiv preprint arXiv:2306.10052*.

Mollick, E. R., & Mollick, L. (2022). New modes of learning enabled by ai chatbots: Three methods and assignments. *Available at SSRN*.

Paul, R., & Elder, L. (2013). Critical Thinking: Intellectual Standards Essential to Reasoning Well Within Every Domain of Human Thought, Part Two. *Journal of developmental education, 37*(1), 32.

Pellegrino, J. H., Margaret. (2012). *Education for life and work: Developing transferable knowledge and skills in the 21st century*: National Academies Press.

Wulandari, R., & Hindrayani, A. (2021). *Measuring Critical Thinking Skills with the RED Model.* Paper presented at the Journal of Physics: Conference Series.